\documentclass[conference]{IEEEtran}
\IEEEoverridecommandlockouts
\usepackage{cite}
\usepackage{amsmath,amssymb,amsfonts}
\usepackage{algorithmic}
\usepackage{graphicx}
\usepackage{textcomp}
\usepackage{xcolor}
\def\BibTeX{{\rm B\kern-.05em{\sc i\kern-.025em b}\kern-.08em
    T\kern-.1667em\lower.7ex\hbox{E}\kern-.125emX}}
\usepackage{algorithm}
\usepackage{algorithmic}

\usepackage{booktabs} % for professional tables
\usepackage{multirow} 

\usepackage{bbm}
\usepackage{amsthm}   % For theorems, definitions, etc.
\usepackage{tcolorbox}
\usepackage{url}

\usepackage{enumitem}

\usepackage{listings}
\usepackage{courier}
\tcbuselibrary{skins,breakable} % <--- THIS FIXES THE ERROR
\tcbset{
    promptbox/.style={
        enhanced,
        colback=white,
        colframe=black!75,
        coltitle=white,
        colbacktitle=black!80,
        fonttitle=\bfseries,
        boxrule=0.5pt,
        arc=2pt,
        left=6pt, right=6pt, top=6pt, bottom=6pt,
        boxsep=4pt,
        breakable,
        title={#1}
    }
}
\lstdefinelanguage{promptlang}{
    basicstyle=\ttfamily\small,
    breaklines=true,
    columns=fullflexible,
    frame=none,
    backgroundcolor=\color{white},
    keywordstyle=\color{blue!60!black}\bfseries,
    morekeywords={insert_node, update_node, move_node, deprecate_node, annotate_node, expand_node, list_children, focus_path, collapse_node},
    moredelim=**[is][\color{red!60!black}]{|}{|},
    moredelim=**[is][\color{purple!60!black}\bfseries]{~}{~},
}

\usepackage{pifont}      % for \ding symbols
\newcommand{\cmark}{\textcolor{green!60!black}{\ding{51}}}
\newcommand{\xmark}{\textcolor{red!70!black}{\ding{55}}}

\theoremstyle{definition}

\usepackage{xspace}

\newcommand{\jasper}{\textsc{Jasper}\xspace}

\newcommand{\model}{\textsc{Grove}\xspace}

\newcommand{\svaEval}{\textsc{SVA-Eval}\xspace}
\newcommand{\svaEvalH}{\textsc{SVA-Eval-Human}\xspace}
\newcommand{\svaEvalM}{\textsc{SVA-Eval-Machine}\xspace}
\newcommand{\llamaThree}{\textsc{LLaMA-3}\xspace}
\newcommand{\oThreeMini}{\textsc{o3-mini}\xspace}
\newcommand{\rag}{\textsc{RAG}\xspace}
\newcommand{\reflexion}{\textsc{Reflexion}\xspace}

\newcommand{\caseRAG}{\textsc{Case-RAG}\xspace}
\newcommand{\reflexionRAG}{\textsc{Ref-RAG}\xspace}
\newcommand{\globalMem}{\textsc{GlobalMem}\xspace}
\newcommand{\reflexionLLM}{\textsc{Ref-seqR}\xspace}

\usepackage[colorinlistoftodos]{todonotes}

\definecolor{applegreen}{rgb}{0.55, 0.71, 0.0}

\begin{document}

\title{Learning to Debug: LLM-Organized Knowledge Trees for Solving RTL Assertion Failures
}

\author{
Yunsheng Bai, Haoxing Ren \\
NVIDIA \\
\texttt{\{yunshengb, haoxingr\}@nvidia.com}
}

\maketitle

\begin{abstract}

Debugging is the dominant cost in modern hardware verification, where assertion failures are among the most frequent and expensive to resolve. While Large Language Models (LLMs) show promise, they often fail to capture the precise, reusable expertise that engineers apply, leading to inaccurate responses. We propose \textbf{\model{}}, a hierarchical knowledge management framework that learns and organizes reusable debugging expertise into an LLM-organized knowledge tree for \emph{solving assertion failures}. \model{} distills debugging knowledge from prior cases and organizes it into a vertical tree of configurable depth, with each node encoding a concise knowledge item and explicit applicability conditions. During training, \model{} uses a parallel, gradient-free loop where an LLM proposes tree modifications as structured JSON edits by learning from the cases. At test time, a budget-aware iterative zoom is performed to navigate the tree, retrieving a small set of applicable knowledge items that guide a base LLM's hypothesis generation and fix proposals. Evaluated on a suite of assertion-failure cases, \model{} delivers consistent gains in pass@1 and pass@5, demonstrating the value of structured knowledge evolution.

\end{abstract}

\begin{IEEEkeywords}
Hardware verification, assertion failures, LLMs, Retrieval-Augmented Generation, knowledge management, RTL debugging
\end{IEEEkeywords}

\section{Introduction} 
\label{sec-intro}

Modern hardware design relies on extensive verification to ensure correctness before fabrication~\cite{foster2020wilson}. Verification at scale produces many failures, yet \emph{debugging}—isolating root causes and crafting RTL fixes—remains the dominant cost. Among failure modes, \emph{assertion failures} are especially costly~\cite{zhou2025insights}, requiring engineers to interpret logs/waveforms, reason about timing/causality, and implement precise RTL edits. LLMs and agents are increasingly used in EDA~\cite{thakur2023autochip,wei2025vericoder}, but naïve prompting or flat retrieval rarely reproduces the \emph{domain-specific, reusable debugging expertise} that practitioners rely on.

This expertise is concrete and action-oriented, targeting patterns that generic models miss, such as correcting spec-to-RTL constant mismatches, fixing off-by-one address predicates, repairing stateful counters, restoring XOR feedback, and zero-extending byte-to-word reads. In prior studies~\cite{zhou2025insights,bai2025fvdebug}, LLM/agent systems often \emph{mislocalize} faults, propose \emph{unstable or over-fitted edits}, and \emph{bloat context} with irrelevant snippets—leading to \emph{inaccurate responses and failed debugging}. Absent a way to \emph{learn, structure, and validate} such reusable knowledge from past cases, performance remains brittle and inefficient.

\begin{figure*}[h]
\centering
\includegraphics[width=0.99\textwidth]{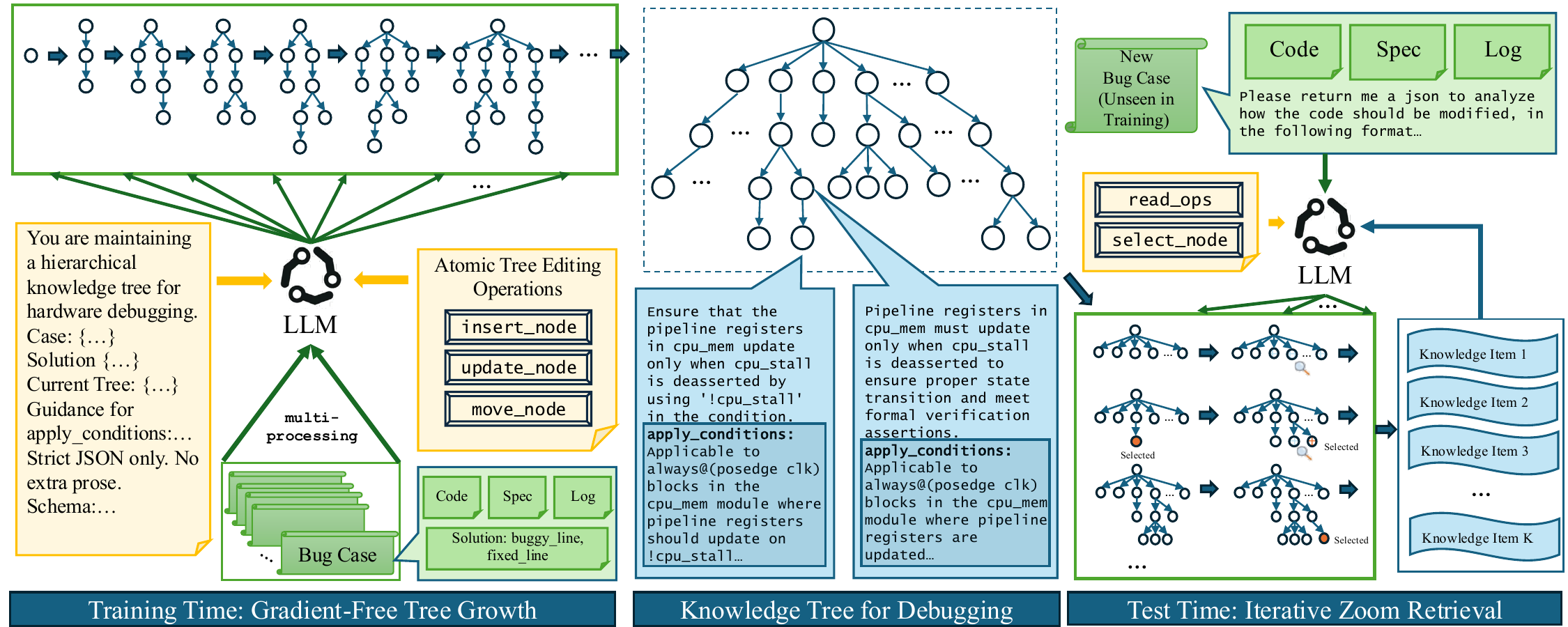}
\vspace{-0.3cm}
\caption{Overview of \model: A knowledge management framework that transforms raw debugging experiences into reusable expertise. Training cases (left) undergo golden-aware reflection to distill validated knowledge items, integrated via atomic JSON edits into an LLM-organized hierarchical knowledge tree (center). Test cases (right) navigate this tree through budgeted snapshot-and-zoom operations to retrieve applicable knowledge, augmenting prompts with targeted debugging guidance. The tree maintains vertical organization (general to specific) and horizontal discipline (via shape guards) to scale across a large amount of training cases while preserving fast, precise retrieval.}
\vspace{-0.3cm}
\label{fig:model_architecture}
\end{figure*}

% Hardware verification at scale routinely surfaces large numbers of failures, yet \emph{debugging}---isolating root causes and crafting fixes---remains the dominant cost~\cite{foster2015trends}. Among failure modes, \emph{assertion failures} are especially common and costly~\cite{zhou2025insights}. Engineers must interpret logs, reason over timing, and translate findings into correct Register-Transfer Level (RTL) edits. While Large Language Models (LLMs) and automated agents show promise in Electronic Design Automation (EDA)~\cite{thakur2023autochip,wei2025vericoder}, na\"ive prompting or general-purpose Retrieval-Augmented Generation (RAG) often fails. These approaches lack the \emph{domain-specific, reusable debugging knowledge} that expert engineers apply. In this context, \textbf{knowledge} refers to the precise, actionable debugging patterns, causal hypotheses, and validated fix strategies (e.g., ``an FSM stuck due to a missing default case'' or ``a spec-to-RTL constant mismatch''). For instance, an agent may repeatedly fail to spot an off-by-one error in a magic number (e.g., `...0505` vs. `...0506`) without prior knowledge that such constant mismatches are a common failure class. Without a system to automatically \emph{learn, organize, and validate} this specialized knowledge, agents remain inefficient and unreliable.

This presents two fundamental challenges. \textbf{First, knowledge acquisition:} How can this latent debugging expertise be automatically learned from unstructured sources like past bug fixes, failure logs, and engineers' experience, moving beyond ad-hoc case matching? \textbf{Second, knowledge organization:} How can this learned knowledge be organized for efficient retrieval and reuse? A flat, unstructured store is demonstrably suboptimal: it pollutes the LLM's context with noisy information, scales poorly, and fails to match the hierarchical way engineers reason (e.g., decomposing an abstract pattern into concrete strategies and fixes).

To solve this, we present \textbf{\model{}} (\textbf{G}overned \textbf{R}etrieval \textbf{O}f \textbf{V}alidated \textbf{E}xpertise), a hierarchical knowledge management framework for solving assertion failures in RTL designs. \model{} distills reusable debugging knowledge from prior cases and organizes it into an \textbf{LLM-organized knowledge tree}. Instead of a fixed hierarchy, \model{} dynamically structures knowledge into a tree of configurable depth. Each node encodes a concise actionable knowledge item, and explicit applicability conditions (\emph{apply\_conditions}) to govern its use.

During training, \model{} runs a parallel, gradient-free learning loop. Each training case produces one or more candidate knowledge items, which are validated by injecting each item into the LLM prompt, generating RTL fixes, and checking them against the assertion using a model-checking tool. Only beneficial, performance-preserving knowledge is committed via \emph{LLM-generated JSON edit scripts}---the LLM autonomously decides where to insert, update, move, or deprecate nodes, thereby organizing and refining the hierarchical tree under structural constraints. At test time, \model{} navigates the tree using an LLM-guided \emph{iterative zoom} on a budgeted snapshot of the tree, filtering candidates by applicability before assembly into a safe, case-specific knowledge set.

% \textbf{This work.} We present \textbf{\model{}} (\textbf{L}earning \textbf{S}elf-evolving \textbf{K}nowledge \textbf{T}rees), a hierarchical knowledge management framework for \emph{solving assertion failures in RTL designs}. \model{} learns reusable debugging knowledge from prior cases and organizes them into a self-evolving tree (category~$\to$~pattern~$\to$~instance). Each node records: (i) a concise actionable knowledge item, (ii) explicit \emph{apply\_conditions} (free-form or structured), and (iii) optional retrieval aids (tags, signatures, code patterns). 

% Training uses a parallel, gradient-free loop to diversify candidates and \emph{validate per-statement} with pass@k; only non-harmful, beneficial knowledge items are committed via strict JSON edit scripts (insert/update/move/deprecate/annotate). At test time, \model{} performs LLM-guided \emph{hierarchical beam search} constrained by applicability and followed by per-item validation to assemble a case-specific, safe rule set.

\textbf{Scope.} We instantiate and evaluate \model{} on \emph{assertion-failure debugging} in RTL. Extension to other debug tasks is conceptually supported by the framework but out of scope for this paper.

Our contributions can be summarized as follows:
\begin{itemize}
  \item An LLM-organized hierarchical knowledge tree with explicit applicability constraints and strict JSON edit operations. To the best of our knowledge, \model{} is the first to combine per-item, tool-validated knowledge acquisition with a governed hierarchical memory tailored to RTL assertion-failure debugging.
  \item A novel LLM-guided zoom procedure for test-time retrieval and a gradient-free, parallel learning workflow for scalable acquisition and organization of hierarchical debugging knowledge.
  % \item A gradient-free, parallel learning workflow with time budgets and early-stopping patience, enabling scalable knowledge acquisition in settings where expensive simulator/prover calls dominate verification costs.
  % \item An LLM-guided zoom procedure for test-time retrieval, which navigates abstraction levels via applicability-aware zoom operations.
  % \item A gradient-free, parallel learning workflow with time budgets and early-stopping patience, enabling scalable knowledge acquisition in settings where expensive simulator/prover calls dominate verification costs.
  \item Comprehensive evaluation on \svaEval, demonstrating \textbf{+12.0\% absolute improvement in pass@1 on \llamaThree{} and +6.6\% on \oThreeMini{}}, with consistent pass@5 gains and ablations validating design choicea.
  % \item Comprehensive evaluation on \svaEval, demonstrating consistent pass@1 and pass@5 improvements over flat RAG and no-memory baselines, with ablations validating design choices.
\end{itemize}

\section{Related Work} 
\label{sec-related}

\subsection{Hardware Debugging and LLMs for EDA}

Hardware debugging has evolved from manual inspection to sophisticated automated approaches. Formal verification~\cite{kern1999formal} and assertion-based verification~\cite{witharana2022survey} established foundational methods, with a recent survey~\cite{wu2024survey} analyzing ML integration across the verification spectrum. ML-driven bug localization~\cite{stracquadanio2024veribug,wu2024knowledge} achieves improvements in identifying faulty regions. Automated repair tools~\cite{ahmad2022cirfix,laeufer2024rtl} address debugging bottlenecks through genetic programming, symbolic execution, and LLM-based approaches. Hardware fuzzing techniques~\cite{laeufer2018rfuzz,jayasena2025fuss} systematically explore state spaces for corner-case bugs. Recent work integrates LLMs into EDA 
workflows~\cite{ho2025verilogcoder,bai2025fvdebug}, 
% workflows~\cite{thakur2023autochip,blocklove2024evaluating,menon2025enhancing,ho2025verilogcoder,bai2025fvdebug}, 
% combining conversational models with tool feedback for iterative design and repair, 
though domain adaptation remains critical.

\subsection{Learning and Accumulating Knowledge}
Valuable expertise for assertion debugging is latent in fixes, logs, and reviews. Beyond static retrieval, \emph{self-learning} methods distill knowledge from feedback via reflection~\cite{shinn2023reflexion, madaan2023self} or persistent, evolving memory stores~\cite{madaan2022memory,zhong2024memorybank,zhang2025survey}. These stores can be dynamic graphs~\cite{xu2025mem}, executable skill libraries~\cite{wang2023voyager}, or architectural augmentations~\cite{wang2023augmenting, langgraph2024memory}. 
\model{} differs by focusing on \emph{governed acquisition} for hardware debugging: knowledge is mined from cases, validated for non-harmfulness, and committed via atomic edits with applicability contracts, building a domain-specific, dynamic tree rather than an unstructured general memory.

\subsection{Organizing and Retrieving Knowledge}
RAG grounds LLMs~\cite{lewis2020retrieval}, but flat case retrieval, common in EDA~\cite{qi2025verirag,wang2025veridebug,qayyum2025llm}, struggles with scale and alignment to hierarchical reasoning. Structure-aware retrieval addresses this using document trees~\cite{zhang2024hierarchical,jiao2025hirag,huang2025retrieval} or knowledge graphs~\cite{edge2024local,jiang2025ras}. This aligns with broader AI trends of deliberate search~\cite{yao2023tree,hogan2021knowledge} and neuro-symbolic composition~\cite{karpas2022mrkl}. Related data system work targets fixed schemas~\cite{freire2025large,lao2025gptuner} rather than evolving domain knowledge. 
In contrast, \model{} \emph{learns} its own dynamic, LLM-organized knowledge tree, combining RAG's grounding with agent memory's adaptivity. 
\section{Methodology}
\label{sec-model}

\subsection{Task Formulation}
\label{sec:task}

We formulate assertion-failure debugging as a two-regime learning and inference task. Figure~\ref{fig:model_architecture} outlines the overall \model framework.

\paragraph{Training cases}
% Following the methodology of prior work~\cite{zhou2025insights}, each training sample consists of (i) a structured problem context comprising RTL code, design specification, and failure logs, and (ii) a \emph{golden solution} in the form of the buggy and fixed code. During training, the LLM is asked to use this information to \emph{distill reusable debugging knowledge}—concise, actionable statements paired with explicit \emph{apply\_conditions}.
Each training sample consists of (i) a structured problem context and (ii) a \emph{golden solution}. Following the methodology of prior work~\cite{zhou2025insights}, the context includes the buggy RTL code, SystemVerilog assertion specifications, and assertion failure logs; the golden solution provides the buggy line(s) and corresponding fix. 

During training, \model{} exposes the golden solution \emph{only for reflection and validation}; the LLM is asked to \emph{distill reusable debugging knowledge}—concise, actionable statements paired with explicit \emph{apply\_conditions} that govern when they apply in natural language (detailed in Section~\ref{sec:train}). 
% Each candidate knowledge item is validated individually using pass@k re-generation; only non-degrading candidates are integrated into the tree via JSON edit scripts (detailed in Section~\ref{sec:train}).

\paragraph{Test cases}
Each test sample provides the same structured context without a golden solution.
\model{} retrieves a \emph{small set} of relevant knowledge items using the budgeted snapshot+zoom protocol described in Section~\ref{sec:snapshot_zoom} (see Algorithm~\ref{alg:grove_zoom}), and \emph{appends the retrieved knowledge} to the test prompt for a base LLM to produce hypotheses and fixes.
This \emph{zoom-then-augment} design yields improved robustness by grounding the LLM's reasoning in validated, case-relevant debugging expertise.

\subsection{Overall Framework}
\label{sec:overall}

\model{} addresses the dual challenges of \emph{knowledge acquisition} and \emph{knowledge organization} by learning reusable, validated debugging knowledge from past cases and maintaining it as a \emph{vertical, LLM-organized knowledge tree}. Each node (knowledge item) carries (i) a concise, actionable statement, and (ii) explicit \emph{apply\_conditions} that govern when the statement should be applied.
% , and (iii) optional retrieval aids (tags, signatures, code patterns, examples).
% The apply\_conditions are surfaced verbatim in the tree views that the LLM sees and describe when the knowledge item should be applied (e.g., specific operators, reset styles, or control constructs).
The \emph{apply\_conditions} are written by the LLM in natural language following a schema that asks for concrete syntactic and semantic cues (e.g., specific operators, module names, signal widths).

Abstraction is \emph{emergent}: upper levels tend to encode more general regularities, while deeper levels capture increasingly specific realizations; this is regulated by these applicability constraints and edit-time validation.
As the tree grows, \model{} enforces simple structural constraints on depth and branching to keep it navigable and efficient; we describe these safeguards in more detail in Sections~\ref{sec:train} and~\ref{sec:shape_scalability}.

\subsection{Handling Growing Tree Size: Snapshot and Zoom}
\label{sec:snapshot_zoom}

\textbf{Motivation.}
A successful deployment of \model{} will accumulate hundreds or thousands of nodes. Exposing the entire tree at once would exceed context budgets and drown useful knowledge in noise. We therefore separate \emph{what is stored} (the full tree) from \emph{what is shown} (a budgeted view), and give the LLM a small set of structured zoom operations to access local detail on demand.

\textbf{Budgeted snapshot.}
\model{} constructs a \emph{budgeted structural snapshot} of the knowledge tree. This snapshot preserves the global shape of the tree while respecting a token budget $B_{\text{snap}}$, and is created by including a representative subset of nodes from each level. For each included node, we render its identifier, a compact summary, and its \emph{apply\_conditions}. Nodes with unexpanded children are marked with ellipsis indicators (e.g., ``\ldots''), signaling to the LLM that zoom operations can reveal additional detail. This design balances \emph{breadth} (global tree shape) with \emph{depth} (local detail on demand).

% \textbf{Budgeted structural snapshot.}
% \model{} constructs a \emph{budgeted structural snapshot} of the knowledge tree that preserves global shape across depths while respecting a token budget $B_{\text{snap}}$. The snapshot is built by traversing the tree level-by-level and including a representative subset of nodes at each level, with configurable per-level caps to maintain proportional coverage. 

% For each included node, we render its identifier, a compact content summary, and (when budget permits) the full \emph{apply\_conditions} text. Nodes with unexpanded children are marked with ellipsis indicators (e.g., ``\ldots''), signaling to the LLM that zoom operations can reveal additional detail. This design balances \emph{breadth} (global tree shape) with \emph{depth} (local detail on demand).

\textbf{JSON-based interaction protocol.}
All LLM--tree interactions use a structured JSON protocol. On each round, \model{} synthesizes a prompt containing the current context $C$, the snapshot $S$, any previously expanded subtrees $V_{\text{details}}$, and a JSON response schema.

% The LLM responds with either \textbf{read operations} or a \textbf{final decision}. Read operations are zoom operations encoded as a \texttt{read\_ops} list containing, for example, \texttt{expand\_node} for subtree expansion or \texttt{list\_children} for immediate descendants; they are executed by a read-only routine under a chunk budget $B_{\text{chunk}}$ to produce additional textual views that are accumulated into $V_{\text{details}}$ across rounds. Final decisions are task-specific: at test time, the LLM returns \texttt{select\_node\_ids} (node IDs for retrieval); during training, it returns a JSON edit script specifying tree modifications (\texttt{insert\_node}, \texttt{update\_node}, \texttt{move\_node}, \texttt{deprecate\_node}) with explicit node references and field updates.

% All LLM--tree interactions use a structured JSON protocol. For each turn, \model{} synthesizes a prompt containing the current context $C$, the snapshot $S$, any previously expanded subtrees, and a JSON response schema. 

The LLM responds with a JSON object that may contain two types of fields: \textbf{(i) read operations}---a \texttt{read\_ops} list specifying zoom actions such as \texttt{expand\_node} (for subtree expansion) or \texttt{list\_children} (for immediate descendants)---and \textbf{(ii) candidate selections}---a \texttt{select\_node\_ids} list marking promising knowledge items. Read operations are executed to generate detailed textual views, which are accumulated across rounds under a per-round chunk budget $B_{\text{chunk}}$.

% The LLM responds with either \textbf{read operations} (zoom operations, encoded as a \texttt{read\_ops} list containing, e.g., \texttt{expand\_node} for subtree expansion or \texttt{list\_children} for immediate descendants) or a \textbf{final decision}. Read operations return textual views accumulated across rounds under a chunk budget $B_{\text{chunk}}$. Final decisions are task-specific: at test time, the LLM returns \texttt{select\_node\_ids} (node IDs for retrieval); during training, it returns a \textbf{JSON edit script} specifying tree modifications (\texttt{insert\_node}, \texttt{update\_node}, \texttt{move\_node}, \texttt{deprecate\_node}) with explicit node references and field updates. 

The semantics of \texttt{select\_node\_ids} are task-specific: at test time, these are the node IDs for retrieval.
During training, \model{} uses an analogous JSON protocol, but instead of \texttt{select\_node\_ids} the LLM returns a \textbf{JSON edit script} specifying tree modifications (\texttt{insert\_node}, \texttt{update\_node}, \texttt{move\_node}, \texttt{deprecate\_node}) with explicit node references and field updates.

% These scripts are structured, machine-readable instructions that specify atomic tree modifications. 

\textbf{Protocol workflow.}
Algorithm~\ref{alg:grove_zoom} formalizes the retrieval protocol, which follows the iterative expansion concept depicted in Figure~\ref{fig:model_architecture}.
The process begins by providing the LLM with an initial, high-level view of the tree: the budgeted snapshot $S$ (Line~\ref{alg:grove_snapshot}), which acts as a navigable map where complex branches are collapsed.
The system then enters a loop for up to $R$ rounds.

In each round, the LLM examines its current view of the tree---the snapshot plus any detailed views requested so far---and can issue \texttt{read\_ops} to ``zoom in'' on specific nodes.
When the LLM requests an expansion, the system generates a more detailed textual representation of that subtree and adds it to the set of detailed views.
On the next iteration, the LLM is presented with an updated view that combines the global snapshot with these expanded regions, allowing it to progressively move from a general overview to specific details.

The loop terminates once the LLM stops requesting further zooms (or the round budget is reached).
At that point, the system aggregates all proposed \texttt{select\_node\_ids}, and returns a final set of node IDs $\hat{\mathcal{K}}$ (Line~\ref{alg:grove_return}).
During training, the same iterative exploration logic applies, but the final JSON object contains an edit script instead of node IDs.

\begin{algorithm}[t]
\caption{\model{} Snapshot+Zoom Retrieval Protocol}
\label{alg:grove_zoom}
\begin{algorithmic}[1]
\REQUIRE Context $C$, tree $T$, snapshot budget $B_{\text{snap}}$, max rounds $R$
\STATE $S \leftarrow \textsc{Snapshot}(T, B_{\text{snap}})$;\quad $V_{\text{details}} \leftarrow \emptyset$;\quad $C_{\text{cand}} \leftarrow \emptyset$ \label{alg:grove_snapshot}
\FOR{$r = 1 \ldots R$}
  \STATE $\text{p} \leftarrow \textsc{ConstructPrompt}(C, S, V_{\text{details}})$
  \STATE $(\texttt{read\_ops}, \texttt{select\_node\_ids}) \leftarrow \textsc{LLM}(\text{p})$
  \IF{\texttt{select\_node\_ids} is non-empty}
    \STATE $C_{\text{cand}} \leftarrow C_{\text{cand}} \cup \texttt{select\_node\_ids}$
  \ENDIF
  \IF{\texttt{read\_ops} is non-empty}
    \STATE $V_{\text{details}} \leftarrow V_{\text{details}} \cup \textsc{ApplyRead}(T,\texttt{read\_ops})$ \label{alg:grove_apply}
  \ELSE 
    \STATE \textbf{break} \label{alg:grove_break} %\COMMENT{No more zoom requested}
  \ENDIF
\ENDFOR
\STATE $\hat{\mathcal{K}} \leftarrow \textsc{Deduplicate}(C_{\text{cand}})$ % \COMMENT{Deduplicate}
\STATE \textbf{return} $\hat{\mathcal{K}}$ \label{alg:grove_return} \COMMENT{Final set of candidate node IDs}
\end{algorithmic}
\end{algorithm}

\subsection{Training-Time Knowledge Acquisition and Validation}
\label{sec:train}

\textbf{Parallel learning workflow.}
To scale knowledge acquisition, \model{} leverages a gradient-free, parallel learning architecture. The system spawns multiple concurrent workers (via multi-processing), where each worker processes a disjoint subset of training cases. All workers share a single underlying knowledge tree data structure. 
% The learning loop for each case proceeds through three stages: reflection, validation, and integration.

\textbf{Golden-aware reflection and tree edits.}
For each training case, \model{} performs a \emph{golden-aware reflection} that inspects the case context alongside the golden fix. Using the snapshot+zoom protocol, the LLM first explores the existing tree to locate relevant regions and potential parents. The reflection step concludes with the LLM proposing a candidate \emph{tree edit}, which it formulates as a \textbf{JSON edit script} as described in Section~\ref{sec:snapshot_zoom}. 

Each script lists one or more operations---\texttt{insert\_node} (add new knowledge), \texttt{update\_node} (modify existing content or metadata), \texttt{move\_node} (restructure hierarchy), or \texttt{deprecate\_node} (mark obsolete)---along with explicit node references (by ID or path) and field updates. This structured, machine-checkable output forces the LLM to formalize its reasoning into executable operations and allows each worker in \model{} to modify a shared tree data structure.

% Training cases are partitioned across multiple parallel workers (separate processes), and each worker independently runs reflection and validation on its own shard of cases, emitting JSON edit scripts against the shared tree to accelerate gradient-free training.

% For each training case, \model{} performs a \emph{golden-aware reflection} that inspects the case context alongside the golden fix. Using the snapshot+zoom protocol, the LLM first explores the existing tree to locate relevant regions and potential parents. The reflection step concludes with the LLM proposing a candidate \emph{tree edit}, which it formulates as a \textbf{JSON edit script} as described in Section~\ref{sec:snapshot_zoom}. This script is then passed to the next stage for validation.

% These edits are expressed as \emph{JSON edit scripts}. 
% Each script lists one or more operations---\texttt{insert\_node} (add new knowledge), \texttt{update\_node} (modify existing content or metadata), \texttt{move\_node} (restructure hierarchy), or \texttt{deprecate\_node} (mark obsolete)---along with explicit node references (by ID or path) and field updates. Scripts are applied under a global write lock, enabling multiple parallel workers each processing a subset of training cases to organize the shared tree without conflicts. This structured output forces the LLM to formalize its reasoning into executable operations and allow each worker of \model{} to modify the underlying shared tree data structure.

\textbf{Per-knowledge-item validation.}
Before modification, the worker evaluates each candidate knowledge item \emph{individually} extracted from the edit scripts. It regenerates solutions for the training case with that single item's statement injected into the prompt. If the item does not degrade pass@k on its originating case,
% (i.e., pass@1 and pass@5 remain at or above baseline), 
it is marked as \emph{non-degrading} and becomes eligible for integration.
% ; candidates that hurt performance are rejected and their associated edit operations are discarded. 
This criterion is chosen over strictly enforcing improvement to enable knowledge generalization and to avoid discarding valid, widely applicable knowledge.
Our current validation scheme uses the originating case, which keeps cost manageable. Extending validation to a separate held-out subset is an interesting future effort.

\textbf{Governed integration.}
Validated scripts are parsed into discrete graph operations and applied atomically to the shared tree under a global write lock. 
This process is termed \emph{governed integration} because \model{} actively enforces structural invariants during execution: schema checks ensure all nodes possess explicit \emph{apply\_conditions}, and a \emph{verticality} constraint forces nodes at level~$l$ to attach only under parents at level~$(l{-}1)$. 
% This active governance prevents the entropy typical of unstructured memories, ensuring the hierarchy maintains a clean taxonomy from general to specific as it grows.
% Eligible items are committed by replaying their JSON edit scripts. Scripts are first checked in a dry-run pass to ensure schema correctness and \emph{verticality} constraints: a node at level~$l$ may only attach under a parent at level~$(l{-}1)$, and all levels must include explicit \emph{apply\_conditions}. 
These checks ensure that abstraction levels progress cleanly from general (upper levels) to specific (lower levels).

% Eligible items are committed by replaying their JSON edit scripts. Scripts are first checked in a dry-run pass to ensure schema correctness and \emph{verticality} constraints: a node at level~$l$ may only attach under a parent at level~$(l{-}1)$, and all levels must include explicit \emph{apply\_conditions}. 
% % Vetted scripts are then applied atomically under a global write lock, so multiple workers can process disjoint training cases in parallel while safely organizing a shared, LLM-organized knowledge tree.
% % Eligible knowledge items are integrated into the hierarchy via \emph{strict JSON edit scripts} (\texttt{insert\_node}, \texttt{update\_node}, \texttt{move\_node}, \texttt{deprecate\_node}), applied atomically under a global write lock, enabling multiple workers to organize the shared tree safely. 
% During each write, \model{} enforces a simple notion of \emph{verticality}: 
% % abstract nodes (levels 1--2) are required to include explicit \emph{apply\_conditions}, and 
% a semantic hierarchy constraint forces a node at level~$l$ to attach only under a parent at level~$(l{-}1)$. 
% These checks ensure that 
% % new knowledge items come with explicit, level-consistent applicability predicates and 
% % that 
% abstraction levels progress cleanly from general (upper levels) to specific (lower levels).

\subsection{Test-Time Knowledge Retrieval}
\label{sec:test}

Embedding-based similarity alone is insufficient for RTL debug retrieval. It ignores \emph{applicability structure}: whether a rule should apply under specific control, temporal, or module contexts. Prior work shows that naïve dense retrieval can pollute context and misalign with downstream tasks when applicability predicates matter~\cite{liu2024lost}, and that tree-organized retrieval can better integrate information across abstraction levels~\cite{sarthi2024raptor}. 

In contrast, \model{} leverages the LLM’s \emph{reasoning} to navigate a budgeted snapshot and issue targeted zoom operations (JSON \texttt{read\_ops}), guided by the learned \emph{apply\_conditions}. This lets the model connect hints present in the test context to rules whose applicability matches, rather than relying on surface-level textual similarity.

At inference, \model{} follows the iterative snapshot+zoom protocol detailed in Algorithm~\ref{alg:grove_zoom}. The LLM progressively accumulates detailed views via \texttt{read\_ops} and marks candidates via \texttt{select\_node\_ids}. 
% The retrieved items (statement + \emph{apply\_conditions}) are appended to the prompt to guide hypotheses and fixes.
The retrieved items are appended as concise statements; \emph{apply\_conditions} guide selection and are surfaced in zoom views.

\subsection{Maintaining Shape and Scalability}
\label{sec:shape_scalability}

To ensure the knowledge tree remains efficient and navigable as it grows, \model{} uses active governance over its shape and records detailed operational telemetry. 
% The system instruments both read operations and write operations, tracking their counts to monitor retrieval patterns and tree evolution over time. 
% This data is used to generate growth plots that summarize per-level node counts and read/write activity, providing insight into the learning process.
Scalability is primarily enforced by \emph{shape guards}, which are strict, configurable checks applied during every write operation. Key parameters include budgets for the maximum number of root-level (level-1) nodes and the maximum children per parent node. By enforcing these constraints at edit time and asking the LLM to re-propose edits when the constraints are violated, \model{} prevents uncontrolled branching and mitigates degradation in retrieval latency and quality as more knowledge items are added.

\subsection{Complexity and Resource Use}
Let $N$ be the number of training cases, $M$ the average number of candidate knowledge items generated per case via golden-aware reflection, $E$ the cost of a single pass@k evaluation (one complete generation and evaluation pass), and $R$ the number of zoom rounds. Training cost is dominated by \emph{per-statement validation} at $O(M\!\cdot\!E)$ per case; edit integration is near-constant per accepted item. Retrieval uses at most $R$ compact prompts, with local expansions touching only targeted subtrees. Parallel workers processing the training data in a multi-processing way yield near-linear speedups until external rate limits or prover/simulator throughput saturate. 
% Shape guards prevent branching from scaling out of control, preserving quality and latency as the tree grows.

\section{Experiments}

We evaluate \model{} on the \svaEval{} benchmark of assertion-failure cases~\cite{zhou2025insights}. Code with data will be released for easy reproduction of our results.
% Unless otherwise noted, all experiments use \llamaThree{} (llama-3.3-70b-instruct) as the base LLM and report pass@1 and pass@5.

\subsection{Dataset}

We adopt \svaEval{}~\cite{zhou2025insights}, which provides specification (\textit{Spec}), buggy code, logs, and a golden fix for each case. We combine the human-curated (\svaEvalH) and LLM-generated (\svaEvalM) splits, then partition the union into 80\% training and 20\% testing, ensuring no module appears in both splits to prevent data leakage. \svaEval{} is produced via tool-validated augmentation: SVAs and bugs are generated and filtered with HDL compilers and formal/symbolic checks to ensure the bug--SVA pair triggers an assertion and the golden fix resolves it~\cite{zhou2025insights}. 
% We refer readers to~\cite{zhou2025insights} for detailed statistics.

\subsection{Baselines}
\label{sec:baselines}

\begin{table}[t]
\centering
\caption{Main evaluation results on the \svaEval{} test set. We report pass@k for all methods across two base LLMs for testing.}
\label{tab:main_results}
\begin{tabular}{l|cc|cc}
\hline
\multirow{2}{*}{\textbf{Method}} & \multicolumn{2}{c|}{\textbf{\llamaThree}} & \multicolumn{2}{c}{\textbf{\oThreeMini}} \\
\cline{2-5}
 & \textbf{Pass@1} & \textbf{Pass@5} & \textbf{Pass@1} & \textbf{Pass@5} \\
\hline
Zero-shot & 0.727 & 0.754 & 0.847 & 0.851 \\
\caseRAG{} & 0.680 & 0.680 & 0.787 & 0.820 \\
\reflexionRAG{} & 0.508 & 0.536 & 0.689	& 0.700 \\
\globalMem{} & 0.732	& 0.765 & 0.836	& 0.847 \\
\reflexionLLM{} & 0.749	& 0.781 & 0.880 &	0.885 \\
\hline
\model{} (\textsc{No-Apply}) & 0.792 & 0.820 & 0.901	& 0.901 \\
\model{} (\textsc{No-Val}) & 0.814 & 0.831 & 0.907	& 0.907 \\
\model{} (\textsc{No-LLM-Ret}) & 0.754 & 0.781 & 0.885	& 0.891 \\
\model{} (\textsc{No-Zoom}) & 0.803 & 0.831 & 0.880 &	0.885 \\
\hline
\textbf{\model{} (Ours)} & \textbf{0.847} & \textbf{0.869} & \textbf{0.913} & \textbf{0.913} \\
\hline
\end{tabular}
\end{table}

We compare \model{} against knowledge management baselines spanning embedding-based retrieval, flat-memory, and LLM-guided selection:
\noindent\textbf{\caseRAG{} (Case-level RAG).}
Each training case is treated as a retrievable document. At test time, we retrieve the top-$k$ most similar cases using dense embeddings and include them in the prompt context.
\noindent\textbf{\reflexionRAG{} (Flat knowledge RAG).}
\reflexion{}~\cite{shinn2023reflexion} distills debugging knowledge from iterative self-reflection into a flat knowledge store. Retrieval follows \caseRAG{} but operates over learned knowledge items rather than full cases.
\noindent\textbf{\globalMem{} (Global memory).}
All \reflexion{} knowledge items are appended to every prompt without retrieval, testing whether unfiltered knowledge access suffices.
\noindent\textbf{\reflexionLLM{} (LLM-based retrieval).}
Inspired by LLM-as-retriever approaches~\cite{asai2024self}, the system chunks the \reflexion{} knowledge store and sequentially prompts the LLM to select relevant items, without \model{}'s hierarchical organization.
For \caseRAG{} and \reflexionRAG{}, we vary $k \in \{1, 2, 4, 8\}$ and report the best result in Table~\ref{tab:main_results} for conciseness.
% We compare \model{} against two knowledge management baselines:

% \noindent\textbf{\rag{} (Vanilla RAG).}
% Each training case (buggy code, spec, logs, solution) is treated as a retrievable document.
% At test time, we retrieve the top-$k$ most similar training cases using dense embeddings (Sentence-BERT~\cite{reimers2019sentence}) and include them in the prompt context.

% \noindent\textbf{\reflexion{}~\cite{shinn2023reflexion}.}
% An iterative self-reflection baseline where the LLM critiques its previous debugging attempts and refines solutions across multiple trials. We then use \rag to retrieve the learned knowledge from the flat knowledge store via \reflexion with different $k$s indicating number of knowledge items retrieved.
% % We allow up to 3 reflection iterations per case, providing the LLM with its prior outputs, failure signals, and a reflection prompt requesting analysis of mistakes.

\subsection{Implementation Details for \model{}}
\label{sec:impl_details}

For \model{}, the knowledge tree is constructed following the workflow described in Section~\ref{sec-model}. The snapshot+zoom protocol is governed by two main budgets: the initial snapshot budget is set to $B_{\text{snap}} = 80000$ tokens, and the per-round chunk budget is $B_{\text{chunk}} = 12000$ tokens.
During testing, we execute at most $R=10$ zoom rounds. 
The tree's structure is managed by shape guards that cap the number of root-level nodes at 216 and the fan-out for any parent node at 144.

\subsection{Evaluation Protocol}

To evaluate the correctness of the generated fixes, we use Cadence \jasper{} (version 2023.12) as the model-checking tool. We apply the patch proposed by each model to the buggy RTL and run the verifier. A fix is considered successful if the modified code compiles and all assertions pass. Consistent with the \svaEval{} benchmark's methodology~\cite{zhou2025insights}, we report the \textbf{pass@1} and \textbf{pass@5} metrics. 
% Pass@1 measures the percentage of problems where the model's first-attempted fix is correct, while pass@5 measures success within the top five attempts. 
Testing is based on two base LLMs: Llama 3.3 70B Instruct~\cite{dubey2024llama} and OpenAI o3-mini~\cite{openai2025o3mini}.

\subsection{Results and Analysis}

Table~\ref{tab:main_results} presents the primary results on \svaEval{}. \model{} delivers the top pass@k across both base LLMs, outperforming zero-shot \llamaThree{} and all non-hierarchical memory baselines.
\caseRAG{} and \reflexionRAG{} provide gains over zero-shot in some settings, confirming that retrieval can help; however, both lag behind \model{}, indicating that naïvely surfacing entire cases or flat memories remains suboptimal.
% With \llamaThree{}, \caseRAG{} and \reflexionRAG{} are uniformly weak, indicating that naïve retrieval of entire, unstructured cases or flat memories can pollute context rather than guiding debugging.
\globalMem{} and \reflexionLLM{} show that simply providing a strong global rulebook or using a flat LLM-based selector is not sufficient. \model{}’s learned, hierarchical knowledge with explicit applicability conditions enables more precise, targeted retrieval.

\subsection{Ablation Study}

% To analyze the contribution of each component in \model{}, we conduct an ablation study on the \svaEval{} test set, 
As shown in Table~\ref{tab:main_results}, removing \emph{LLM-guided retrieval} (\textsc{No-LLM-Ret})—i.e., replacing \emph{snapshot+zoom} with embedding-based similarity search—causes the largest degradation in pass@1, indicating that semantic similarity alone cannot capture the nuanced applicability conditions needed for effective selection. 

Removing \emph{applicability conditions} (\textsc{No-Apply}) from the tree also harms performance, confirming that explicit, natural-language predicates are essential for precise routing; and dropping the \emph{iterative zoom protocol} (\textsc{No-Zoom}) underperforms the budgeted, progressive exploration used by \model{}. 

Finally, removing \emph{per-statement validation} (\textsc{No-Val}) further reduces accuracy. Together, these ablations confirm that \model{}'s gains arise from the synergy of validated knowledge, hierarchical organization, explicit applicability, and LLM-guided retrieval.

\subsection{Retrieval Quality Study}
\label{sec:retrieval-quality}
Beyond ablations, we additionally evaluate retrieval quality with an LLM-as-judge~\cite{es2024ragas}, which sees the case context and a method’s retrieved items and returns a calibrated \emph{helpfulness} score $[0,1]$. We also judge whether a method’s final answer is \emph{relevant} (\textbf{AnsRel}) and \emph{supported} (\textbf{AnsSup}) by its retrieved items. Table~\ref{tab:retrieval_quality} shows \model{} retrieves more helpful items, and its answers are more often relevant and supported than non-hierarchical baselines. 

\begin{table}
\centering
\small
\caption{Retrieval quality (LLM-as-judge) on \llamaThree{}. Higher is better.}
\label{tab:retrieval_quality}
\begin{tabular}{lccc}
\hline
\textbf{Method} & \textbf{Helpfulness} & \textbf{AnsRel} & \textbf{AnsSup} \\
\hline
\caseRAG{} & 0.221 & 0.419 & 0.193 \\
\reflexionRAG{} & 0.315 & 0.471 & 0.281 \\
\globalMem{} & 0.279 & 0.440 & 0.215 \\
\reflexionLLM{} & 0.441 & 0.587 & 0.368 \\
\textbf{\model{}} & \textbf{0.626} & \textbf{0.711} & \textbf{0.539} \\
\hline
\end{tabular}
\end{table}

\subsection{Characteristics of the Learned Knowledge Tree}

We analyze \model{}'s learning process and the resulting tree structure. Our gradient-free, parallel workflow is efficient: with 8 parallel workers, the full knowledge tree was learned from the 80\% training split (732 cases) in under 24 hours.

\begin{figure}[h]
\centering
\includegraphics[width=1.0\columnwidth]{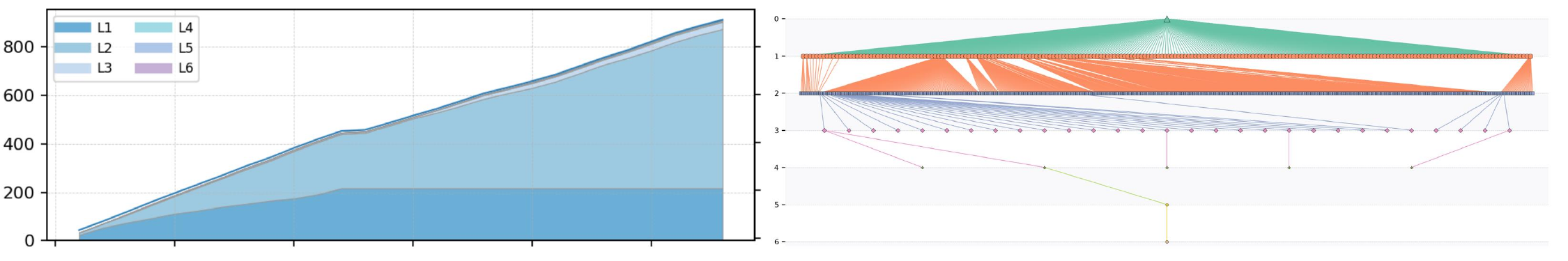}
\caption{Knowledge tree growth and structure. \textbf{Left:} Stacked area of nodes per level across training steps; the upper envelope gives the cumulative node count, demonstrating steady, continuous learning. \textbf{Right:} Final tree layout showing hierarchical organization across six levels.}
\label{fig:tree_growth}
\end{figure}

Figure~\ref{fig:tree_growth} visualizes the tree's growth dynamics. 
We observe steady, monotonic accumulation at every level, indicating that \model{} continuously distills and integrates new expertise rather than oscillating or collapsing categories. 
The right panel provides a structural view of the final tree, illustrating the hierarchical organization across six levels. The majority of knowledge resides at L2 and L3, representing core knowledge.

The \textbf{\emph{emergent}} hierarchy—organized by the LLM rather than hand-designed—arises from governed learning with per-statement validation, explicit applicability conditions, and shape guards that maintain depth and fan-out discipline.

% We analyze the scalability of \model{}'s learning process and the resulting tree structure. Our gradient-free, parallel workflow is highly efficient: with 8 parallel workers, the full knowledge tree was learned from the 80\% training split (732 cases) in under 24 hours.

% Figure~\ref{fig:tree_growth} visualizes the tree’s growth dynamics. 
% % The left panel is a stacked area chart of node counts per level (L1--L4) over training steps; the upper envelope corresponds to the cumulative number of nodes. 
% We observe steady, monotonic accumulation at every level, indicating that \model{} continuously distills and integrates new expertise rather than oscillating or collapsing categories. 
% The right panel provides a structural view of the final 336-node tree, illustrating the hierarchical organization across four levels.
% Overall, the tree is shallow but expressive and remains vertically governed rather than flat. This structure directly enables efficient \emph{snapshot+zoom} retrieval.
% % ,
% % : the LLM first makes a few high-level routing decisions at L1/L2, then zooms into rich, validated rules at L3/L4 with minimal search.
% The \textbf{\emph{emergent}} hierarchy—rather than being manually designed—arose naturally from governed learning, where per-rule validation and explicit applicability conditions enforce coherent organization.

% \begin{table}[h]
% \centering
% \caption{Final Knowledge Tree Statistics}
% \label{tab:tree_stats}
% \begin{tabular}{lc}
% \hline
% \textbf{Statistic} & \textbf{Value} \\
% \hline
% Total Nodes & 336 \\
% Max Depth & 4 \\
% Nodes at Level 1 & 24 \\
% Nodes at Level 2 & 126 \\
% Nodes at Level 3 & 172 \\
% Nodes at Level 4 & 13 \\
% \hline
% \end{tabular}
% \end{table}

\subsection{Data Efficiency and Computational Cost}
% \subsection{Effect of Training Set Size}
\label{sec:train-ratio}

\begin{figure}[h]
  \centering
  \includegraphics[width=0.9\columnwidth]{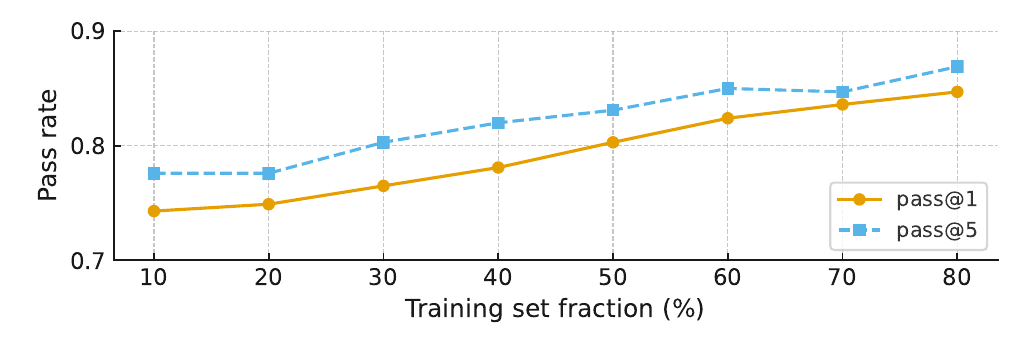}
  \caption{
  Effect of training set size for \model{} with \llamaThree{}.
  }
  \label{fig:train_ratio}
\end{figure}

To study robustness under limited training data, we vary the fraction of \svaEval{} training cases used to train \model{}, keeping the test set fixed (Figure~\ref{fig:train_ratio}).
Both pass@1 and pass@5 improve steadily as more data are available, indicating that \model{} remains effective under data scarcity and continues to benefit from additional cases.
% We note \model{}'s iterative retrieval consumes approximately $3\times$ more tokens than flat RAG baselines.
% However, this investment inference-time compute yields \textit{data efficiency} gains.
% In practical debugging scenarios, this higher latency is often negligible compared to the hours of engineering time saved by a correct first-pass fix.
Across all methods, end-to-end latency per case including inference and evaluation is dominated by \jasper{} runs; the additional LLM calls introduced by \model{} keep total time within the same order of magnitude as the strongest baselines.

% Across all methods, end-to-end latency per case---including LLM inference and evaluation---is dominated by \jasper{} runs; the additional LLM calls introduced by \model{} add moderate overhead but keep total time within the same order of magnitude as the strongest baselines.

% \subsection{Retrieval-Time Efficiency}
% \label{sec:runtime}

% We report average end-to-end wall-clock time per test case inference and validation in Figure~\ref{fig:runtime} for \llamaThree. 
% \globalMem{} is fastest
% % among knowledge-based methods since it bypasses retrieval entirely, 
% but suffers in accuracy. 
% Embedding-based retrieval (\caseRAG{}, \reflexionRAG{}) incurs moderate latency, while \reflexionLLM{} is slowest due to sequential LLM scanning over flat knowledge.
% % \model{} completes in 64.94 seconds—faster than \reflexionLLM{} while achieving the best accuracy.
% % In practice, \model{} strikes a favorable accuracy–latency trade-off: it avoids the steep runtime growth of high-$k$ retrieval while delivering the best pass@1/5.
% Absolute wall-clock times depend on LLM endpoint latency and throughput, but the key insight holds: \model{}'s structured retrieval adds moderate overhead while yielding the largest accuracy gains. Future work could explore techniques to improve the efficiency, e.g. adaptive retrieval (learning when knowledge lookup is beneficial in an agentic system), knowledge compression, etc.

% \begin{figure}[h]
% \centering
% \includegraphics[width=0.9\columnwidth]{runtime.pdf}
% \caption{Average runtime per test case (seconds) across 183 test cases. 
% % ``Ref'' denotes ``\reflexion{}''.
% % Bars (left to right): Zero-shot (3.67), RAG $k{=}1,2,4,8$ (25.56, 45.13, 57.76, 58.56), Reflexion+RAG $k{=}1,2,4,8$ (52.62, 58.56, 74.00, 81.11), and \model{} (64.94). A dashed horizontal grid aids comparison.
% }
% \label{fig:runtime}
% \end{figure}

\subsection{Case Study: Training-Time Knowledge Acquisition}
\label{sec:case-train}

Figure~\ref{fig:train_case} illustrates \model{}'s knowledge acquisition
process. Given a training case where the register width
mismatches the specification, the LLM performs golden-aware reflection and
proposes a JSON edit script to insert a new level-2 knowledge item under an
existing parent. 
The proposed item then undergoes per-statement validation (pass@k unchanged on the originating case)
before atomic integration via the structured edit operation.

% Figure~\ref{fig:train_case} illustrates \model{}'s knowledge acquisition process. Given a training case where a shift register incorrectly shifts data, the LLM performs golden-aware reflection and proposes a JSON edit script to insert a new level-3 knowledge item under an existing parent about direct assignments. 
% The proposed knowledge item undergoes per-statement validation (pass@1 remains 1.0 on the originating case).
% , and is then atomically integrated into the tree via the structured edit operation.

\begin{figure}[t]
\centering
\scriptsize
\begin{minipage}[t]{0.48\columnwidth}
\vspace{0pt}
\raggedright
\setlength{\tabcolsep}{2pt}
\renewcommand{\arraystretch}{1.08}
\begin{tabular}{|p{0.96\linewidth}|}
\hline
\textbf{Training Case: dataGenerator} \\
\hline
\textbf{Assertion} \\
\texttt{@(posedge clock) disable iff(!nReset)} \\
\texttt{!testModeFlag |=\textgreater{} (adcData == \$past(adc\_databus));} \\
\hline
\textbf{Spec (excerpt)} \\
\#\# 2. Module Interface \\
\#\#\# Inputs \\
- **adc\_databus** (10-bit): Input bus carrying data from an ADC. \\
\hline
\textbf{Buggy RTL (excerpt)} \\
\texttt{reg [8:0] adcData;} \\
\texttt{assign dataOut = testModeFlag ? testData : adcData;} \\
\texttt{always @(posedge clock, \ldots} \\
% \texttt{always @(posedge clock, negedge nReset) begin \ldots end} \\
\hline
\textbf{Golden fix} \\
\texttt{reg [9:0] adcData;} \\
\hline
\end{tabular}
\vspace{0.2em}

\textbf{(a)} Training case with golden solution.
\end{minipage}\hfill
\begin{minipage}[t]{0.48\columnwidth}
\vspace{0pt}
\raggedright
\setlength{\tabcolsep}{2pt}
\renewcommand{\arraystretch}{1.08}
\begin{tabular}{|p{0.96\linewidth}|}
\hline
\textbf{LLM-proposed JSON edit (excerpt)} \\
\hline
\ttfamily\footnotesize
"ops": [\{\\
\ \ "type": "insert\_node",\\
\ \ "parent\_ref": \{"id": \ldots \},\\
\ \ "node": \{\\
\ \ \ \ "level": 2,\\
\ \ \ \ "title": "DataGenerator Width Mismatch in ADC Data Capture",\\
\ \ \ \ "knowledge\_statement": "Ensure that capture registers use the exact width specified for their input buses to prevent data capture assertion failures.",\\
% \ \ \ \ "knowledge\_statement": "Ensure that register widths exactly match the specification to prevent data capture assertion failures.",\\
% \ \ \ \ "apply\_conditions": "Applicable when an ADC capture register should be 10-bit per spec but is declared narrower (e.g., 9-bit). Look for 'reg [8:0] adcData;' when the spec and properties expect 10 bits."\\
\ \ \ \ "apply\_conditions": "Applicable when an ADC capture register is declared narrower than the specified bus width (e.g., reg [8:0] for a 10-bit bus). Look for width mismatches between RTL registers and spec/properties."\\
\ \ \} \}]  \\
\hline
\end{tabular}
\vspace{0.2em}

\textbf{(b)} Proposed edit and validation.
\end{minipage}
\vspace{0.5em}
\caption{Training-time knowledge acquisition. (a) Golden-aware reflection
on \texttt{dataGenerator}. (b) LLM proposes
inserting a level-2 knowledge item.}
\label{fig:train_case}
\end{figure}

\subsection{Case Study: Test-Time Knowledge Retrieval}
\label{sec:case-test}

Figure~\ref{fig:test_case_125} presents a case study on an Avalon-interface FIFO counter with memory-mapped address decoding. The bug is subtle: the RTL uses bitwise AND (\texttt{\&}) instead of a zero-extending operation, causing the \texttt{readdata} to be all zeros instead of a zero-extended value. As shown, \model{}-\llamaThree retrieves a highly relevant knowledge item about zero-extension. 
Guided by this, it produces a correct fix, while the baseline, lacking this context, misdiagnoses the problem.

\begin{figure}[t]
\centering
\scriptsize
\begin{minipage}[t]{0.46\columnwidth}
\vspace{0pt}
\raggedright
\textbf{Parent knowledge item (L1).}
Ensure that both read ports in register-file modules conditionally output zero for address~0.  \\
\enspace \enspace \textit{apply\_conditions}: This rule now serves as a grouping node for interface assertions related to register file and memory modules \ldots \\[0.25em]
\textbf{Children (L2):}
\begin{itemize}[leftmargin=1.8em,itemsep=1pt,parsep=0pt]
  % \item[\texttt{|--}] Ensure \texttt{read\_mux\_out} is assigned based on \texttt{(address == 0)} to propagate the correct SPI input on address~0 \dots
  \item[\texttt{|--}] Ensure any write operation sets register~0 to zero 
  % regardless of input data 
  \dots
  \item[\texttt{|--}] \textbf{Retrieved knowledge item ($\bigstar$).}\\
  \emph{``Read multiplexer assignments must use (address == 0) to route valid data, ensuring that \texttt{data\_out} is zero-extended for non-zero addresses per specification.''} \\
  \enspace \enspace \textit{apply\_conditions}: When constructing the read multiplexer, check the address condition so that the produced data matches \ldots \\
  \item[\texttt{|--}] Replace the incorrect bit slice assignment \texttt{[4:1]} with \texttt{[3:0]} to match the assertion requirement \dots
  % \item[\texttt{|--}] Replace incorrect bit-slice \texttt{[4:1]} with \texttt{[3:0]} in related status registers \dots
  % \item[\texttt{|--}] Assign \texttt{readdata} using OR (\texttt{|}) instead of AND (\texttt{\&}) to zero-pad upper bits \dots
  % \item[\texttt{\textbackslash--}] Ensure pipeline registers propagate full-width data signals without unintended bit slicing \dots
\end{itemize}

\textbf{(a)} Knowledge tree context around the retrieved knowledge item ($\bigstar$).
\end{minipage}\hfill
\begin{minipage}[t]{0.50\columnwidth}
\vspace{0pt}
\raggedright
\setlength{\tabcolsep}{2pt}
\renewcommand{\arraystretch}{1.05}
\begin{tabular}{|p{0.96\linewidth}|}
\hline
\textbf{Assertion} \\
\texttt{address == 0 |=> readdata[5:0] == \$past(in\_port)} \\
\texttt{\ \ \&\& readdata[31:6] == 0;} \\
\hline
\textbf{Buggy RTL (line 32)} \\
\texttt{readdata <= \{32'b0 \& read\_mux\_out\};} \\
\textit{Bitwise AND masks \texttt{read\_mux\_out} to 0} \\
\hline
\textbf{Golden fix (from dataset)} \\
\texttt{readdata <= \{32'b0 | read\_mux\_out\};} \\
\textit{Bitwise OR with 32'b0 zero-extends the 6-bit value to 32 bits} \\
\hline
\textbf{Baseline fix (zero-shot)} \\
\texttt{assign read\_mux\_out = (address == 0) ? data\_in : 6'b0; \ldots} \\ % \texttt{\ \ $\rightarrow$ (address == 0) ? data\_in : 6'b0;} \\
{\color{red}\xmark}\ \textit{Jasper: FAILED (wrong location)} \\
\hline
\textbf{\model{} fix (with knowledge)} \\
\texttt{readdata <= \{26'b0, read\_mux\_out\};} \\
{\color{green}\cmark}\ \textit{Jasper: PASSED. In this module, both this fix and the golden form a 32-bit zero-extension of the 6-bit \texttt{read\_mux\_out}.} \\
% \hline
% \textbf{Result} \\
% Pass@1: Baseline 0.0 vs.\ \model{} 1.0 \\
\hline
\end{tabular}

\textbf{(b)} Fix comparison.
\end{minipage}
\caption{Test-time knowledge retrieval. (a) Learned knowledge visualization. (b) Effect on fix generation.
% on the \texttt{ulight\_fifo\_counter\_rx\_fifo} design.
}
\label{fig:test_case_125}
\end{figure}

\section{Conclusion and Future Work}
\label{sec-conc}

% We introduced \model{}, a hierarchical knowledge management framework that learns self-evolving trees for assertion failure debugging in RTL designs. Experimental results demonstrate significant pass@1 and pass@5 improvements over flat RAG and no-memory baselines. Future work could extend this structured knowledge paradigm to other EDA stages such as test generation, coverage closure, and design optimization, where hierarchical expertise organization may similarly benefit automated debugging and design workflows.

We presented \model{}, a hierarchical knowledge management framework for RTL assertion-failure debugging. Our work demonstrates that an LLM can effectively maintain and organize a dynamic, hierarchical knowledge base. This gradient-free learning process allows the knowledge tree to be organized by the LLM as a structured, symbolic store, integrating validated expertise via atomic edits rather than costly backpropagation. Experimental results demonstrate \model{}'s effectiveness, showing pass@1 and pass@5 improvements over baselines.

Future work could extend \model{} to other debugging tasks (e.g., value mismatches, timing violations) and broader EDA tasks (e.g., RTL generation, test synthesis, coverage closure). Key research challenges include: distilling robust knowledge in complex design scenarios, enabling continual learning for evolving codebases, and further reducing retrieval latency. Addressing these challenges would unlock more capable and efficient agents across design automation.

% Future work could extend \model{} to other debugging modalities (value mismatches, timing violations) and broader EDA tasks (RTL generation, test synthesis, coverage closure). Key challenges include learning from limited training data, distilling knowledge in complex scenarios where root causes are subtle, and enabling cross-task knowledge transfer where expertise from one EDA stage accelerates learning for related tasks. Addressing these could unlock more robust, data-efficient automation across hardware design workflows.

% \begin{IEEEkeywords}
% component, formatting, style, styling, insert
% \end{IEEEkeywords}

{
% \small
\bibliographystyle{plain}
\bibliography{bibliography}
}

\appendices

\section{Preliminaries and Task Setup}
\label{app:background}

RTL verification is the critical process of ensuring that a hardware implementation matches its architectural specification. In practice, this is primarily achieved through two complementary methodologies: dynamic simulation, which applies test vectors to the design and observes behavior over time~\cite{bergeron2000writing}, and Formal Property Verification (FPV), which mathematically checks that the design satisfies specified properties under all legal input conditions~\cite{clarke2018model}. SystemVerilog Assertions (SVAs)~\cite{sutherland2006systemverilog} are the standard way to encode design intent in both flows, capturing temporal relationships such as protocol handshakes and invariants. For instance, the property \texttt{request |-> \#\#[1:2] grant} mandates that whenever \texttt{request} is asserted, \texttt{grant} must follow within one to two clock cycles. Such monitors provide localized observability into signal relationships and protocol violations.

While SVAs may generally reside directly in the RTL or in a testbench, the \svaEval benchmark~\cite{zhou2025insights} utilized in this work adopts a specific structure. Each case is packaged with a natural-language specification, the RTL code, and a set of SVAs associated with the design. In this context, we operate under the standard debugging assumption that the provided SVAs correctly reflect the intended design behavior, while the RTL implementation contains the faults. To ensure a fair comparison, we reuse these exact artifacts—specification, RTL, and logs—without modification.

When an assertion fails, it indicates a divergence between the implementation and the specification. In FPV, the tool produces a counter-example (CEX) trace—a precise sequence of states leading from a reset to the failure. In this work, we utilize \jasper (Cadence JasperGold) to verify assertions. While the original dataset generation in \cite{zhou2025insights} utilized the open-source \textsc{SymbiYosys} tool ~\cite{symbiyosys}, we adopt \jasper due to its trusted status and widespread adoption in industrial verification flows. To ensure consistent evaluation, we rigorously verified that all buggy RTL samples generate a valid CEX and all golden repairs result in proven properties under \jasper, preventing mismatches caused by solver differences.

We note that recent agentic approaches, such as \textsc{FVDebug}~\cite{bai2025fvdebug}, have begun incorporating waveform modalities directly into the debugging loop. However, our work distinguishes itself by focusing on the \textit{learning} and \textit{retrieval} of debugging knowledge to enhance textual prompting, rather than multi-modal waveform analysis. Our approach is orthogonal to such agentic systems; the knowledge retrieval mechanism proposed here effectively serves as an atomic reasoning operation that could be plugged into waveform-aware agents.

Finally, while this work focuses specifically on assertion failures, hardware debugging encompasses broader scopes, such as analyzing value mismatches in data paths or diagnosing simulation hangs (deadlocks). While the core principles of \model apply in theory to these broader scenarios, we leave their exploration to future work and focus here on the task of resolving SVA failures.

\section{Extended Comparison with Related Work}
\label{sec-comparison}

\begin{table*}[t]
\centering
\caption{Comparison of \model{} with related knowledge management and retrieval approaches. \cmark{} indicates the feature is present; \xmark{} indicates absence; $\sim$ indicates partial or indirect support.}
\label{tab:comparison}

\resizebox{0.83\textwidth}{!}{% trick!

\begin{tabular}{@{}lcccccc@{}}
\toprule
\textbf{Method} & \textbf{Hierarchical} & \textbf{Explicit} & \textbf{Tool-Validated} & \textbf{Dynamic} & \textbf{LLM-Guided} & \textbf{Hardware} \\
& \textbf{Organization} & \textbf{Applicability} & \textbf{Knowledge/Fixes} & \textbf{Learning} & \textbf{Retrieval} & \textbf{Specific} \\
\midrule
\multicolumn{7}{l}{\emph{Our baselines on \svaEval{}}} \\
\caseRAG{} & \xmark & \xmark & \xmark & \xmark & \xmark & \cmark \\
\reflexionRAG{} & \xmark & \xmark & \xmark & \cmark & \xmark & \cmark \\
\globalMem{} & \xmark & \xmark & \xmark & \cmark & \xmark & \cmark \\
\reflexionLLM{} & \xmark & \xmark & \xmark & \cmark & $\sim$ & \cmark \\
\midrule
\multicolumn{7}{l}{\emph{Hardware debugging / EDA-specific systems}} \\
CirFix~\cite{ahmad2022cirfix} & \xmark & \xmark & \cmark & \xmark & \xmark & \cmark \\
RTL-Repair~\cite{laeufer2024rtl} & \xmark & \xmark & \cmark & \xmark & \xmark & \cmark \\
VeriBug~\cite{stracquadanio2024veribug} & \xmark & \xmark & \xmark & \xmark & \xmark & \cmark \\
VeriRAG~\cite{qi2025verirag} & \xmark & \xmark & $\sim$ & \xmark & \xmark & \cmark \\
VeriDebug~\cite{wang2025veridebug} & \xmark & \xmark & $\sim$ & \xmark & \xmark & \cmark \\
\textsc{FVDebug}~\cite{bai2025fvdebug} & \xmark & \xmark & $\sim$ & \xmark & \xmark & \cmark \\
\midrule
\multicolumn{7}{l}{\emph{Domain-agnostic knowledge, memory, and \rag{} frameworks}} \\
RAPTOR~\cite{sarthi2024raptor} & \cmark & \xmark & \xmark & \xmark & \xmark & \xmark \\
Self-RAG~\cite{asai2024self} & \xmark & \xmark & \xmark & \cmark & $\sim$ & \xmark \\
Voyager~\cite{wang2023voyager} & \xmark & \xmark & $\sim$ & \cmark & \xmark & \xmark \\
GraphRAG~\cite{edge2024local} & \cmark & \xmark & \xmark & \xmark & \xmark & \xmark \\
A-MEM~\cite{xu2025mem} & \cmark & \xmark & \xmark & \cmark & \xmark & \xmark \\
CAM~\cite{li2025cam} & \cmark & \xmark & \xmark & \cmark & \cmark & \xmark \\
\midrule
\model{} (Ours) & \cmark & \cmark & \cmark & \cmark & \cmark & \cmark \\
\bottomrule
\end{tabular}%

}

\end{table*}

Table~\ref{tab:comparison} provides a comparison of \model{} against related approaches. We group methods into three categories: (i) our baselines (not including ablation study versions of \model{} as in Table~\ref{tab:main_results}), (ii) hardware debugging / EDA-specific systems, and (iii) domain-agnostic knowledge, memory, and \rag{} frameworks. We compare along six axes central to debugging-oriented knowledge systems:

\begin{enumerate}[leftmargin=*]
    \item \textbf{Hierarchical Organization}: whether the method maintains an explicitly multi-level or structured store rather than a flat memory.
    \item \textbf{Explicit Applicability}: whether individual knowledge items are equipped with explicit predicates or conditions describing \emph{when} they should apply.
    \item \textbf{Tool-Validated Knowledge/Fixes}: whether external tools (e.g., provers, simulators) are used as part of the acceptance/rejection criterion for knowledge or fixes.
    \item \textbf{Dynamic Learning}: whether the system accumulates information across runs (beyond static pre-training), e.g., via reflection, memory updates, or continual learning.
    \item \textbf{LLM-Guided Retrieval}: whether an LLM actively steers retrieval (e.g., via selection, zoom, or routing), as opposed to fixed similarity search alone.
    \item \textbf{Hardware Specific}: whether the method is explicitly designed for RTL / EDA tasks (as opposed to being domain-agnostic).
\end{enumerate}

\subsection{Method Comparison}

\textbf{\caseRAG{}.}
Our simplest baseline retrieves the top-$k$ most similar training cases via dense embeddings and appends them to the prompt. It is hardware-specific (applied to \svaEval{}) but has no dynamic learning, hierarchy, or LLM-guided selection---retrieval is purely similarity-based.

\textbf{\reflexionRAG{}, \globalMem{}, and \reflexionLLM{}.}
These baselines adapt \reflexion{}-style self-learning~\cite{shinn2023reflexion} to \svaEval{}. All three distill knowledge from iterative reflection into a flat store, hence they support dynamic learning but lack hierarchical organization. \globalMem{} dumps all learned items into the prompt without retrieval (LLM-Guided is thus inapplicable in a strict sense; we mark \xmark{} since no selection occurs). \reflexionLLM{} uses an LLM to sequentially filter items from the flat store, which we mark as $\sim$ for LLM-Guided Retrieval---it is LLM-driven but operates over an unstructured memory without tree navigation or zoom.

\textbf{CirFix and RTL-Repair.}
These automated repair tools~\cite{ahmad2022cirfix,laeufer2024rtl} generate candidate fixes and use formal/symbolic tools to validate them, hence \cmark{} for Tool-Validated. However, they do not accumulate a persistent knowledge base: each repair session starts fresh. They lack hierarchy, applicability predicates, dynamic learning, and LLM-guided retrieval.

\textbf{VeriBug.}
VeriBug~\cite{stracquadanio2024veribug} uses attention-based ML for bug localization. It is hardware-specific but does not use external tool validation for its predictions (hence \xmark{}), does not accumulate knowledge across designs, and retrieval is implicit in the model rather than LLM-guided.

\textbf{VeriRAG, VeriDebug, and \textsc{FVDebug}.}
These recent LLM-based debugging systems~\cite{qi2025verirag,wang2025veridebug,bai2025fvdebug} interact with EDA tools (coverage analyzers, provers, waveform viewers) during their workflows, which justifies $\sim$ for Tool-Validated---they validate \emph{outputs} or use tool feedback to guide search, but they do not maintain a persistent knowledge store whose entries are individually tool-checked before integration. None of these systems organize knowledge hierarchically or attach explicit applicability conditions. They do not perform dynamic cross-session learning (each debugging session is independent), and retrieval (where applicable) is embedding-based rather than LLM-navigated.

\textbf{RAPTOR.}
RAPTOR~\cite{sarthi2024raptor} constructs a static tree of document summaries via recursive LLM-based abstraction. It is hierarchical (\cmark{}) but the tree is built once from a fixed corpus---there is no dynamic learning or governed evolution. Retrieval is embedding-based over the tree nodes; the LLM is used for \emph{constructing} summaries, not for \emph{steering} retrieval at inference time, hence \xmark{} for LLM-Guided Retrieval.

\textbf{Self-RAG.}
Self-RAG~\cite{asai2024self} trains an LLM to decide when to retrieve and to critique its own outputs, enabling a form of LLM-guided retrieval ($\sim$) and dynamic self-improvement (\cmark{}). However, it operates over flat document stores without hierarchy or explicit applicability conditions, and it is domain-agnostic.

\textbf{Voyager.}
Voyager~\cite{wang2023voyager} builds a skill library for embodied agents, where skills are validated via environment feedback ($\sim$ for Tool-Validated). It supports dynamic learning but the skill library is flat (no hierarchy), lacks explicit applicability predicates, and retrieval is embedding-based.

\textbf{GraphRAG.}
GraphRAG~\cite{edge2024local} organizes documents into a knowledge graph with community summaries, providing hierarchical structure (\cmark{}). However, the graph is constructed statically from a corpus, there is no dynamic learning, and retrieval is primarily graph-traversal and embedding-based rather than LLM-steered.

% \textbf{A-MEM and CAM.}
% A-MEM~\cite{xu2025mem} and CAM~\cite{li2025cam} represent advanced agentic memories. CAM, in particular, builds a constructivist, hierarchical schema via online integration. While they share the goal of dynamic, structured memory (\cmark{}), they lack tool-based validation and do not attach explicit natural-language applicability conditions to nodes---selection relies on embedding similarity or structural adjacency.

\textbf{A-MEM.}
A-MEM and related agentic memory systems~\cite{xu2025mem,zhong2024memorybank} maintain richer structures than flat buffers (e.g., graphs or clustered memories), justifying \cmark{} for hierarchical organization and dynamic learning. However, they lack explicit applicability conditions and hardware specialization, and retrieval remains primarily similarity-based.

\textbf{CAM.}
CAM~\cite{li2025cam} introduces a constructivist agentic memory for long-form reading comprehension, building an overlapping hierarchical memory via incremental clustering and online integration. At inference time, it adaptively explores this structure to activate query-relevant information. However, CAM does not attach explicit applicability predicates to knowledge items---selection is driven by embedding similarity and structural adjacency rather than natural-language conditions. It also lacks tool validation, and is domain-agnostic (targeting reading comprehension rather than hardware debugging).
% CAM~\cite{li2025cam} proposes a constructivist agentic memory for long-form reading comprehension, grounded in Piaget-inspired notions of structured schemata, flexible assimilation, and dynamic accommodation. It builds an overlapping, hierarchical memory via incremental clustering and online integration, and adaptively explores this structure at inference time to activate query-relevant information. However, CAM does not attach explicit applicability predicates to knowledge items---selection is driven by embedding similarity and structural adjacency rather than natural-language conditions. It also lacks tool validation; cluster summaries are generated by LLMs without external verification. CAM is domain-agnostic (targeting reading comprehension) and does not address hardware-specific debugging. 

\textbf{\model{} (Ours).}
Taken together, these comparisons position \model{} as, to our knowledge, the first \emph{governed hierarchical memory} that combines tool-validated, applicability-aware knowledge with LLM-guided retrieval for RTL assertion-failure debugging.

\subsection{Insights and Discussion}

The comparison in Table~\ref{tab:comparison} highlights that debugging occupies a distinctive niche within the broader landscape of agentic memory and \rag{} systems. Unlike open-ended generation tasks, debugging typically operates in a ``pattern-rich but not pattern-free'' regime: real-world RTL bugs are diverse, yet they tend to cluster around recurring failure modes such as handshake mismatches, off-by-one counters, unsafe resets, or incomplete assumptions. 
% \model{} leverages this structure by extracting, validating, and organizing debugging patterns into a governed, hierarchical store.

\textbf{When transferability holds—and when it breaks.}
Debugging tends to succeed when buggy implementations represent \emph{bounded deviations} from a plausible golden reference. In these settings, learned patterns—whether provided by humans or extracted from prior cases—have clear applicability. However, two illustrative classes of scenarios challenge this assumption:

\begin{enumerate}[leftmargin=*]
    \item \emph{Severely degraded or poorly structured RTL.}  
    If a junior engineer or a weakly trained LLM produces RTL that deviates drastically from any reasonable implementation, the ``buggy'' code is no longer a small perturbation of a ``golden'' reference. In such cases, learned debugging patterns offer little guidance: the task shifts from \emph{repair} to \emph{rewrite}.

    \item \emph{Complex or underspecified designs.} At the other extreme, consider scenarios where multiple valid implementations exist for a given specification, or where macro-architectural choices are not uniquely determined by the spec. An LLM may not know which variant the original designer intended, and such design-intent knowledge is difficult to acquire from prior bug-fix pairs alone. Similarly, cutting-edge designs may have incomplete specifications where the ``golden'' implementation is itself uncertain. In these cases, the challenge is ambiguity in the target---knowledge acquisition presupposes a stable ground truth, and when that truth is contested or multi-valued, learning becomes harder. Extending \model{} to such complex scenarios remains an avenue for future work.
\end{enumerate}

These scenarios delineate the \emph{effective radius} of knowledge-driven debugging: the approach excels when defects are localized perturbations of stable design intent and can degrade or even fail when the divergence grows too large.

\textbf{Why a tree, and not a general graph?}
Our choice is motivated by constraints unique to RTL debugging:

\begin{itemize}[leftmargin=*]
    \item Debugging knowledge is typically \emph{hierarchically scoped}: broad invariants (e.g., reset safety) govern narrower patterns (e.g., asynchronous-reset ordering). A tree mirrors how engineers conceptualize debugging heuristics.

    \item \emph{Governed growth.} General graphs grow dense and cyclic, making applicability hard to control. A tree with shape guards enforces vertical abstraction boundaries and keeps retrieval efficient.
    
    \item Debugging requires \emph{diagnostic narrowing}, not just associative linking. Graphs excel at lateral connections, but trees excel at \emph{exclusion}. A debugger naturally moves from symptoms (root) → subsystem category → specific root cause (leaf). Trees enforce this monotonic refinement. The snapshot+zoom protocol leverages this structure: the LLM prunes irrelevant branches early via high-level applicability checks.
\end{itemize}

We do not claim trees are universally optimal. For tasks where knowledge is inherently multi-relational (e.g., entity-centric QA) or where flexible assimilation across overlapping themes is paramount, graph or cluster structures may be preferable. For RTL debugging, reasoning typically progresses from general symptoms toward specific root causes---even when multiple relevant nodes are retrieved, the underlying abstraction levels remain well-ordered. Trees naturally encode this vertical discipline while still allowing the retrieval of multiple nodes across different branches.

\textbf{Outlook.}
Debugging is a domain where formal rigor, engineering heuristics, and pattern-based learning naturally intersect. We believe the core insight of \model{}---that debugging expertise can be captured, validated, and organized into a governed hierarchical store---will remain relevant as agents mature. Looking ahead, we hypothesize that a critical next frontier is developing agents that understand the \emph{boundaries of their own competence}. Such systems should be capable of distinguishing when a new case falls within the manifold of their prior training---where stored patterns are reliable---and when it represents a novel deviation requiring first-principles reasoning or human intervention. Hardware verification, with its high cost of failure and rich diversity of failure modes, offers a particularly compelling proving ground for building these increasingly \emph{self-aware} agents---and we hope \model{} serves as an invitation for the community to explore this exciting direction.

\section{Implementation Details and Practical Considerations}
\label{app:impl_details}

\subsection{Tree Storage Format and Scalability}
\label{app:tree_storage}

Our reference implementation of \model{} represents the knowledge tree as an in-memory, process-shared data structure backed by Python's \texttt{multiprocessing.Manager}~\cite{pythondocs}. The \texttt{Manager} class provides a way to create shared objects---such as dictionaries and lists---that can be safely accessed by multiple concurrent processes. Under the hood, a manager spawns a separate server process that holds the actual data; worker processes communicate with this server via proxies, ensuring that updates from one worker are immediately visible to others. This shared-memory mechanism is essential for our parallel learning workflow, where multiple workers process disjoint subsets of training cases while modifying a single shared tree.

The tree itself is anchored by a \emph{root node} (with identifier \texttt{"root"} and level~0), whose children form the top-level categories of the knowledge hierarchy. All nodes---including the root---reside in a single manager-backed dictionary that maps node identifiers to node objects, enabling $O(1)$ lookup by ID. Traversal from the root downward is performed by following each node's \texttt{children\_ids} list, which is also manager-backed to support safe concurrent updates.

Each knowledge item is represented as a Python dictionary with member variables for the \emph{knowledge statement}, \emph{apply\_conditions}, level, parent identifier, and an ordered list of children identifiers. Node identifiers are generated deterministically by hashing the node's content string using MD5.

\textbf{Concurrency control.}
Read-mostly operations (e.g., snapshot construction and zoom views) acquire a re-entrant read lock (\texttt{manager.RLock()}), whereas \emph{edit scripts} from the LLM are applied under a global write lock (\texttt{manager.Lock()})~\cite{pythondocs}. These synchronization primitives ensure that snapshot generation can proceed concurrently across workers, while tree modifications are serialized to maintain structural invariants.

\textbf{Checkpointing.}
During training, the tree is periodically serialized into a plain Python dictionary and saved as a pickled file, accompanied by a human-readable textual dump. These checkpoints serve multiple purposes: they enable \emph{resumption} of interrupted training runs, support \emph{incremental learning} by loading a previously trained tree and continuing to add knowledge from new cases, and allow a trained tree to be loaded at test time.

\textbf{Merging checkpoints.}
It is possible to combine multiple checkpoint files into a single unified tree---for example, consolidating knowledge learned from different data partitions. In the future, more sophisticated merging strategies could detect and resolve semantic duplicates. This process is conceptually analogous to \emph{model merging}~\cite{yang2024model} or \emph{knowledge distillation}~\cite{hinton2015distilling} in neural networks, but operates on an explicit, interpretable symbolic store.

\textbf{Scaling considerations.}
The current design is sufficient for our experimental scale. However, as the number of knowledge items grows toward industrial scales---e.g., millions of nodes spanning many designs---several bottlenecks would emerge: memory pressure from holding the entire tree in RAM; lock contention when many workers attempt concurrent modifications; serialization overhead for large checkpoint files; and LLM API rate limits that throttle parallel workers during training.

\textbf{Decoupling storage backends.}
To support larger knowledge trees, a natural path forward is to treat the in-memory representation as a \emph{logical interface} and decouple it from the physical storage backend. The tree APIs can be reimplemented on top of external stores without changing the higher-level learning or retrieval logic.

One option is a graph database such as \textsc{Neo4j}~\cite{neo4j}, where nodes become graph vertices and parent--child links become directed edges. Depth-limited traversals and per-level caps can be expressed directly as Cypher queries.

An alternative is a relational backend with linked SQL tables: a \texttt{nodes} table storing per-item content and metadata, and an \texttt{edges} table storing parent--child relationships. Hierarchical queries can be implemented using recursive common table expressions~\cite{burzanska2009pushing}.

In both settings, rarely accessed subtrees can be migrated to cold storage while keeping high-traffic regions in memory. Sharding by top-level category offers a straightforward way to distribute a very large tree across machines.

Crucially, these backend changes affect only the storage layer, not the learning semantics of \model{}. The gradient-free parallel learning loop, per-statement validation, and governed integration via JSON edit scripts continue to operate against the same abstract tree API. 
% This separation of concerns enables a smooth transition to more scalable infrastructure as the knowledge tree grows. 
Future work can explore compression techniques to maintain the symbolic store's compactness while preserving its interpretability.

\subsection{Budgeted Snapshot Construction}
\label{app:snapshot_construction}

Algorithm~\ref{alg:grove_zoom} in the main text invokes a \textsc{Snapshot} routine to produce a budgeted view of the tree. This appendix details how that snapshot is constructed to balance \emph{coverage} (showing enough of the tree's structure) with \emph{compactness} (respecting the token budget $B_{\text{snap}}$).

\begin{algorithm}[t]
\caption{Budgeted Snapshot Construction}
\label{alg:snapshot}
\begin{algorithmic}[1]
\REQUIRE Tree $T$ with root node, token budget $B_{\text{snap}}$
\ENSURE Textual snapshot $S$ respecting $B_{\text{snap}}$
\STATE $\textit{lines} \gets [\,]$;\quad $\textit{used} \gets 0$
\STATE $\textit{top} \gets \text{children of root in } T$
\FOR{each $n \in \textit{top}$} %\COMMENT{Phase 1: all level-1 nodes}
    \STATE $\ell \gets \textsc{RenderNode}(n)$
    \STATE Append $\ell$ to $\textit{lines}$
    \IF{$n$ has children} 
      \STATE Append ellipsis marker to $\textit{lines}$ 
    \ENDIF
\ENDFOR
\STATE $\textit{used} \gets \textsc{Tokens}(\textit{lines})$
\STATE $\textit{frontier} \gets$ all children of nodes in $\textit{top}$
\WHILE{$\textit{frontier} \neq \emptyset$} % \COMMENT{Phase 2: budget-aware BFS}
    \STATE Take and remove a node $n$ from $\textit{frontier}$
    \STATE $\ell \gets \textsc{RenderNode}(n)$
    \IF{$\textit{used} + \textsc{Tokens}([\ell]) > B_{\text{snap}}$} \label{alg:snap_budget_check}
        \STATE Append ``\texttt{... [budget exhausted; use read\_ops to expand]}'' to $\textit{lines}$;\quad 
        \STATE  \textbf{break}
    \ENDIF
    \STATE Append $\ell$ to $\textit{lines}$;\quad $\textit{used} \gets \textit{used} + \textsc{Tokens}([\ell])$
    \IF{$n$ has children}
        \STATE Append ellipsis marker to $\textit{lines}$
        \STATE Add all children of $n$ to $\textit{frontier}$
    \ENDIF
\ENDWHILE
\STATE \textbf{return} $\textsc{Join}(\textit{lines})$
\end{algorithmic}
\end{algorithm}

\textbf{Design goals.}
The snapshot must satisfy three properties: (i)~preserve the global shape of the tree so the LLM can reason about where to zoom; (ii)~include a representative subset of nodes from deeper levels where budget permits; and (iii)~terminate cleanly at the budget boundary without mid-line truncation, which would confuse the LLM parser.

\textbf{Two-phase construction.}
We construct the snapshot in two phases, summarized in Algorithm~\ref{alg:snapshot}. In the first phase, we unconditionally include \emph{all} level-1 nodes (direct children of the root). These top-level categories define the coarse structure of the knowledge hierarchy and are essential for navigation; omitting any would hide entire branches from the LLM. For each level-1 node, we render its identifier, a compact content summary, and its \emph{apply\_conditions}, followed by an ellipsis marker if it has unexpanded children.

In the second phase, we perform a budget-aware breadth-first traversal starting from level-2 nodes. Before appending each node’s text to the output, we check whether doing so would exceed the remaining token budget (Line~\ref{alg:snap_budget_check}); if so, we emit a budget-exhausted marker and terminate. Nodes with unexpanded children are marked with ellipsis indicators (e.g., ``\texttt{...}''), signaling to the LLM that zoom operations can reveal additional detail in those branches.

\textbf{Token estimation and rendering.}
Token counts are estimated using the \texttt{tiktoken} library~\cite{tiktoken} with the \texttt{cl100k\_base} encoding. Each included node is rendered as a single line containing: (i)~its unique identifier (for use in subsequent \texttt{read\_ops}); (ii)~its level in the hierarchy; (iii)~its content text; and (iv)~its \emph{apply\_conditions}. Indentation reflects depth, providing visual structure.

In practice, the guarantee that all level-1 nodes can be shown without exceeding $B_{\text{snap}}$ relies on the shape guards described in Section~\ref{sec:shape_scalability}, which cap the number of root-level categories during training. If, in a different configuration, substantially more top-level nodes were allowed, one would either need to (i)~instruct the LLM (via edit-time prompts) to consolidate or move nodes so that the number of root children remains within a manageable bound, or (ii)~apply a similar budgeted policy to level~1 itself, showing only a subset of top-level nodes per snapshot and relying on successive zoom rounds to explore the remainder.

\subsection{Training-Time Knowledge Acquisition Protocol}
\label{app:training_protocol}

While Algorithm~\ref{alg:grove_zoom} describes the retrieval process at test time, the training loop involves a more complex interaction where the LLM acts as an active \emph{editor} of the knowledge tree. This process is formalized in Algorithm~\ref{alg:grove_train}.

\textbf{Golden-aware reflection.}
Unlike at test time, the training prompt includes the \emph{golden solution} (buggy code and correct fix). The LLM is instructed to reflect on the discrepancy between the buggy and fixed code (line~\ref{alg:grove_train_llm} of Algorithm~\ref{alg:grove_train}), identify the underlying principle or pattern that was violated, and then explore the current tree to propose read and write operations.

We note that reflection on golden solutions is one of several possible knowledge acquisition strategies. In scenarios where agent trajectories are available---e.g., logs from multi-step debugging sessions with tool calls, intermediate hypotheses, and eventual fixes---one could adopt contrastive learning approaches that compare successful versus failed reasoning paths, or distill knowledge from the differences between trajectories that led to correct fixes and those that did not. For instance, given two agent runs on the same bug where one succeeds and one fails, one could extract knowledge by prompting an LLM to identify what the successful trajectory did differently. Multi-agent debate or self-play could also generate diverse perspectives. In this work, we adopt the simpler golden-aware reflection paradigm as it directly leverages the ground-truth fixes available in our benchmark. Exploring richer knowledge acquisition strategies for more complex, open-ended debugging tasks remains an avenue for future work.

\textbf{Edit proposal and JSON parsing.}
Instead of returning \texttt{select\_node\_ids}, the training-time LLM produces a \emph{JSON edit script} containing a list of operations under an \texttt{"ops"} key. The system parses JSON by searching for fenced code blocks (\texttt{```json ... ```}) or bare JSON objects/arrays in the LLM output, retrying up to 10 times if parsing fails. Each operation is a dictionary specifying its \texttt{"type"} and relevant fields. We support the following operation types (with more details found in Section~\ref{app:prompts-training}):

\begin{itemize}[leftmargin=*, itemsep=2pt, topsep=2pt]
    \item \texttt{insert\_node}: Creates a new knowledge item. Required fields include \texttt{"parent\_ref} (either \texttt{\{"id": "node\_xxx"\}} or \texttt{\{"path": ["Category Title", ...]\}} to specify the parent node), and \texttt{"node"} containing \texttt{"level"}, \texttt{"title"}, \texttt{"knowledge\_statement"}, and \texttt{"apply\_conditions"}.

    \item \texttt{update\_node}: Modifies an existing node's metadata. Required fields are \texttt{"node\_ref"} (containing the target node's \texttt{"id"}) and \texttt{"fields"} (a dictionary of attributes to overwrite).

    \item \texttt{move\_node}: Relocates a node to a different parent while preserving its identifier and descendants. Required fields are \texttt{"node\_ref"} and \texttt{"new\_parent\_ref"}. The system detaches the node from its original parent's children list and attaches it under the new parent.

    \item \texttt{deprecate\_node}: Marks a node as deprecated.
\end{itemize}

All operations reference nodes by their unique identifiers (visible in the snapshot and zoom views). Shape guards enforce budgets on the number of top-level categories and patterns per category, rejecting insertions that would exceed these limits. Semantic hierarchy constraints ensure that a level-$k$ node is always a direct child of a level-$(k{-}1)$ node.

\begin{algorithm}[t]
\caption{Training-Time Knowledge Acquisition Protocol}
\label{alg:grove_train}
\begin{algorithmic}[1]
\REQUIRE Context $C$, golden solution $G$, tree $T$, snapshot budget $B_{\text{snap}}$, max rounds $R$
\STATE $S \leftarrow \textsc{Snapshot}(T, B_{\text{snap}})$;\quad $V_{\text{details}} \leftarrow \emptyset$
\FOR{$r = 1 \ldots R$}
  \STATE $p \leftarrow \textsc{ConstructTrainingPrompt}(C, G, S, V_{\text{details}})$
  \STATE $(\texttt{read\_ops}, \texttt{edit\_script}) \leftarrow \textsc{LLM}(p)$\label{alg:grove_train_llm}
  \IF{\texttt{read\_ops} is non-empty}
    \STATE $V_{\text{details}} \leftarrow V_{\text{details}} \cup \textsc{ApplyRead}(T,\texttt{read\_ops})$\label{alg:grove_train_read}
  \ENDIF
  \IF{\texttt{edit\_script} is non-empty}
    \FOR{each \texttt{op} in \texttt{edit\_script}}
      \STATE $K_{\text{cand}} \leftarrow \textsc{ExtractKnowledge}(\texttt{op})$
      \STATE $\text{valid} \leftarrow \textsc{Validate}(C, K_{\text{cand}})$ \COMMENT{Per-item validation}
      \IF{$\text{valid}$}
        \STATE $\textsc{ApplyEdit}(T, \texttt{op})$ \COMMENT{Atomic update under global write lock}
      \ENDIF
    \ENDFOR
    \STATE \textbf{break} \COMMENT{Finished editing for this case}
  \ENDIF
\ENDFOR
\end{algorithmic}
\end{algorithm}

\section{Appendix: Prompt Details}
\label{app:prompts}

This appendix provides the exact prompt templates used in GROVE. We split the \textit{Knowledge Acquisition Prompt} used during training into three parts for clarity.

\subsection{Training}
\label{app:prompts-training}

\begin{tcolorbox}[promptbox={Prompt 1 (Part 1/3): Context \& Reflection}]
\begin{lstlisting}[language=promptlang]
You are maintaining a hierarchical knowledge tree for hardware debugging. You may first request READ ops to zoom the tree, then propose WRITE ops.

Case Context:
- Issue: |ISSUE_NAME|
- Design: |DESIGN_NAME|

Original Prompt:
--- BEGIN PROMPT ---
|FULL_CASE_CONTEXT: Specification, Buggy RTL, Assertion Log|
--- END PROMPT ---

Reflection (for training; do not generate fixes/hypotheses):
--- BEGIN GOLDEN ---
|GOLDEN_SOLUTION_AND_EXPLANATION|
--- END GOLDEN ---
\end{lstlisting}
\end{tcolorbox}

\begin{tcolorbox}[promptbox={Prompt 1 (Part 2/3): Tree State \& Task Instructions}]
\begin{lstlisting}[language=promptlang]
Current Tree (budgeted skeleton view; includes node ids):
--- BEGIN TREE ---
|BUDGETED_TREE_SNAPSHOT|
--- END TREE ---

Optional Expanded Views (if any):
|EXPANDED_SUBTREES|

Task:
- Build a vertical hierarchy.
- Abstraction policy (dynamic):
  - Increase specificity with depth.
  - Do not hard-code role labels; focus on precise, actionable rules.
  - Prefer minimal necessary depth; merge/move to reduce redundancy.

- Prefer inserting under the RIGHT parent using parent_ref by path or id.
- Allowed ops: ~insert_node~ (for NEW content), ~move_node~ (restructure), ~update_node~ (metadata only), ~deprecate_node~.
- DO NOT use update_node to add new rules. New rules must be insert_node.

SEMANTIC HIERARCHY (ENFORCED):
- A node at level k MUST be a direct child of a node at level k-1.
\end{lstlisting}
\end{tcolorbox}

\begin{tcolorbox}[promptbox={Prompt 1 (Part 3/3): Schema \& Examples}]
\begin{lstlisting}[language=promptlang]
- For each new or modified node, include ~apply_conditions~ as NATURAL LANGUAGE. Be specific and deterministic; reference operators, reset style, widths, control constructs, or short code tokens.
- Include tags and signatures when helpful for retrieval.

REQUIRED: include apply_conditions (string or object) for insert_node/update_node.

Return a STRICT JSON edit script with a small number of operations. No extra prose. Schema:
{
  "version": 1,
  "read_ops": [ ... ],
  "ops": [ ... write ops as below ... ]
}

Write ops example:
{
  "version": 1,
  "ops": [
    {
      "type": "~insert_node~",
      "parent_ref": {"path": ["CATEGORY TITLE"]},
      "node": {
        "level": 2,
        "title": "Pattern title",
        "knowledge_statement": "One-sentence actionable rule",
        "apply_conditions": "Natural language description of applicability; concise or verbose as needed."
      }
    },
    {
      "type": "~move_node~",
      "node_ref": {"id": "node_abc"},
      "new_parent_ref": {"id": "node_parent"}
    },
    {
      "type": "~update_node~",
      "node_ref": {"id": "node_xyz"},
      "fields": {
        "apply_conditions": "Update the applicability in natural language if content changes.",
      }
    },
    {
      "type": "~deprecate_node~",
      "node_ref": {"id": "node_old"},
      "reason": "duplicate or harmful"
    }
  ]
}
\end{lstlisting}
\end{tcolorbox}

\subsection{Iterative Zoom Retrieval (Testing)}
At test time, GROVE uses a \textit{Retrieval Prompt} that presents a budgeted skeleton of the tree and allows the LLM to issue \texttt{read\_ops} to explore relevant branches before selecting the final set of knowledge nodes.

\begin{tcolorbox}[promptbox={Prompt 2: Iterative Zoom Retrieval}]
\begin{lstlisting}[language=promptlang]
Select relevant knowledge nodes from the tree for this case.

You may first request READ ops to zoom the tree (expand_node/list_children), then return select_node_ids.
Return STRICT JSON with keys ~select_node_ids~ (when final) and optional ~read_ops~.

READ ops (use to explore before selecting):
- ~expand_node~: {"type":"expand_node", "node_ref":{"id":"node_x"}, "depth":2} -- show a budgeted subtree under a node.
- ~list_children~: {"type":"list_children", "node_ref":{"id":"node_x"}} -- list children.

Context: |HYPOTHESIS_OR_PROBLEM_DESCRIPTION|

TREE (budgeted skeleton):
|BUDGETED_TREE_SNAPSHOT|

Optional Expanded Views (if any):
|ACCUMULATED_EXPANSIONS_FROM_PREVIOUS_ROUNDS|

JSON schema:
{
  "select_node_ids": ["node_xxx", "node_yyy"],
  "read_ops": []
}
\end{lstlisting}
\end{tcolorbox}

\subsection{Retrieval Quality Evaluation (LLM-as-Judge)}
\label{app:retrieval-eval-prompt}

To evaluate retrieval quality (Section~\ref{sec:retrieval-quality}), we use an LLM-as-judge that scores the helpfulness of retrieved knowledge items and evaluates whether the generated answer is relevant to and supported by those items.

\begin{tcolorbox}[promptbox={Prompt 3: Retrieval Quality Evaluation}]
\begin{lstlisting}[language=promptlang]
[SYSTEM]
You are a STRICT and CONSERVATIVE judge for retrieval-augmented RTL debugging.
- Output ONLY JSON (no prose).
- Calibrate scores conservatively; avoid 1.0 unless conditions are fully met.
- Prefer fewer items if relevance is uncertain.

[USER]
Task: Judge retrieved knowledge items for a hardware assertion failure debugging case.

Case: |ISSUE_NAME|
Problem description:
|PROBLEM_DESCRIPTION|

Retrieved knowledge items (index:numbered):
|RETRIEVED_ITEMS_LIST|

If an answer/fix was proposed, it's shown below (optional):
|PROPOSED_FIX_OR_NONE|

Output JSON schema:
{
  "~retrieval_helpfulness~": 0.0,   // float in [0,1]: overall usefulness of the retrieved items to fix this case
  "~answer_relevance~": 0,          // 0/1: does the answer/hypothesis address the problem?
  "~answer_supported~": 0           // 0/1: does the answer rely on or get supported by the retrieved items?
}

Rules:
- Return JSON ONLY. No text outside JSON.
- If no answer was provided, set answer_relevance=0 and answer_supported=0.
- Calibration rubric for retrieval_helpfulness (cap values accordingly):
  * 1.0 ONLY if at least one kept item explicitly matches the failing predicate/signals/operators in this case and would directly justify the fix.
  * <=0.7 if items are plausible but not explicitly tied to failing signals/predicates/operators.
  * <=0.4 if tangential (weak connection).
  * 0.0 if irrelevant.
\end{lstlisting}
\end{tcolorbox}

\end{document}